\documentclass{article} 
\usepackage{iclr2026_conference,times}
\iclrfinalcopy

\usepackage{amsmath,amsfonts,bm}

\def\eqref#1{equation~\ref{#1}}

\def\1{\bm{1}}

\DeclareMathAlphabet{\mathsfit}{\encodingdefault}{\sfdefault}{m}{sl}
\SetMathAlphabet{\mathsfit}{bold}{\encodingdefault}{\sfdefault}{bx}{n}

\usepackage{hyperref}
\usepackage{url}

\usepackage{multirow}
\usepackage{booktabs, colortbl, xcolor, graphicx}
\usepackage{amsmath,amssymb,amsfonts}
\usepackage{graphicx}
\usepackage{textcomp}
\usepackage{xcolor}
\usepackage{algorithm}
\usepackage{algorithmicx}
\usepackage{algpseudocode}
\usepackage{pgfplots}      
\usepackage{caption}      
\usepackage{subcaption} 
\usepackage{amsmath}
\def\BibTeX{{\rm B\kern-.05em{\sc i\kern-.025em b}\kern-.08em
    T\kern-.1667em\lower.7ex\hbox{E}\kern-.125emX}}

\title{Exploring Cross-model Neuronal Correlations in the Context of Predicting Model Performance and Generalizability}

\author{
Haniyeh Ehsani Oskouie$^{1}$, Sajjad Ghiasvand$^{2}$, Lionel Levine$^{1}$, Majid Sarrafzadeh$^{1}$\\
\\[-1em]
$^{1}$University of California, Los Angeles \quad
$^{2}$University of California, Santa Barbara\\
\texttt{haniyeh@cs.ucla.edu}
}

\begin{document}

\maketitle

\begin{abstract}
\normalsize
As Artificial Intelligence (AI) models are increasingly integrated into critical systems, the need for a robust framework to establish the trustworthiness of AI is increasingly paramount. While collaborative efforts have established conceptual foundations for such a framework, there remains a significant gap in developing concrete, technically robust methods for assessing AI model quality and performance. This paper introduces a novel approach for assessing a newly trained model's performance based on another known model by calculating the correlation between neural networks. The proposed method evaluates correlations by determining if, for each neuron in one network, there exists a neuron in the other network that produces similar output. This approach has implications for memory efficiency, allowing for the use of smaller networks when high correlation exists between networks of different sizes.  
On ImageNet-pretrained ResNets, DenseNets, and EfficientNets, partial layer comparisons recover intuitive architectural affinities, indicating that the procedure scales with reasonable approximations. These results support representational alignment as a lightweight compatibility check that complements standard accuracy and calibration, enabling early external validation of new models.


\end{abstract}

\section{Introduction}

Artificial intelligence (AI) has moved from supporting roles to the core of what many societies regard as essential services, from healthcare to public safety \cite{guilherme2019ai}. As these systems influence increasingly consequential decisions, their validation can no longer rely on ad hoc checks or narrow task accuracy alone. Robust, transparent approaches are needed to establish whether a model behaves reliably under real-world conditions and evolving inputs \cite{batarseh2021survey,sujan2023validation}. In parallel, regulators and licensing bodies are signaling that AI systems will need oversight and audit practices commensurate with other high-stakes technologies \cite{clark2019regulatory,o2019legal}.

Today's validation toolkits remain heavily dependent on developer-controlled ingredients—training and validation data, simulation harnesses, and expert judgments—which are invaluable yet inherently internal \cite{myllyaho2021systematic}. Post-market monitoring adds an important safety net, but by definition it observes failures only after deployment \cite{myllyaho2021systematic}. What is missing is a complementary pathway for \emph{external} and \emph{independent} assessment that does not require privileged access to training data, internal agents, or proprietary evaluation suites.

Motivated by this gap, we explore whether \emph{representational alignment} between a candidate model and a well-audited reference model can serve as a practical, computable indicator of trustworthiness. Intuitively, if two networks encode inputs similarly across their layers, they may exhibit related strengths and failure modes; conversely, stark representational divergence can be an early warning that a new model is operating outside familiar behavior regimes.

Prior representation comparison methods have taken different approaches to this problem. Canonical correlation techniques such as SVCCA and PWCCA compare learned subspaces across layers and training runs \cite{raghu2017svcca,morcos2018pwcca}, while Centered Kernel Alignment offers a scalable similarity measure that correlates with transfer performance \cite{kornblith2019cka}. Representational Similarity Analysis, meanwhile, has linked deep network features to neural and human similarity judgments \cite{kriegeskorte2008rsa}. These methods primarily target layer- or subspace-level diagnostics. Our approach is complementary and intentionally minimalist: rather than subspace alignment, we compute a direct neuron-level correlation score with a lightweight depth penalty, designed for external auditing without access to training data. This connects conceptually to observations that independently trained models often learn similar features, including non-robust ones susceptible to adversarial transfer \cite{papernot2016transfer,ilyas2019adversarial}.

Concretely, given a small probe dataset used \emph{only} to elicit activations, we compute, for each neuron in one model, the best-correlated neuron in the other (using absolute correlation), apply a depth-aware penalty to respect architectural hierarchy, and then average scores bidirectionally. The result is a single scalar in $[0,1]$ that increases with representational alignment. For modern architectures, we estimate this efficiently by comparing corresponding layers or stages and, where necessary, subsampling neurons. Crucially, the procedure requires no access to training data or internal training artifacts—supporting independent evaluation alongside existing, developer-centric methods \cite{batarseh2021survey,sujan2023validation,myllyaho2021systematic}.\footnote{%
Throughout, we use ``training data independent'' to mean that our procedure does not require access to a model's \emph{training} data or labels, nor to proprietary evaluation suites.
A small, unlabeled probe set is still used solely to elicit activations; therefore the method is not strictly ``data free'', and estimates can vary with the probe distribution.
}

We evaluate this metric on ImageNet-pretrained ResNets, DenseNets, and EfficientNets. Partial-layer analyses recover intuitive architectural affinities (for example, depth-adjacent variants aligning most), indicating that the score scales to contemporary models with tractable approximations.

\paragraph{Contributions.}
\begin{enumerate}
    \item We propose a simple, symmetric neuronal-correlation metric with a layer-aware penalty that can be computed without access to training data.
    \item We demonstrate a tractable partial-correlation procedure on large ImageNet models that recovers plausible architectural relationships, supporting the metric's utility at scale.
\end{enumerate}

Together, these results position cross-model neuronal correlation as a lightweight \emph{compatibility check}: a model-agnostic indicator linking representational alignment to empirical robustness, designed to complement existing validation practices and emerging regulatory expectations \cite{clark2019regulatory,o2019legal}.



\section{Proposed Algorithm}

This section describes the proposed method for quantifying the similarity between two trained neural networks based on their internal representations. The goal is to produce a symmetric, data-independent measure that reflects how similarly two models process and encode inputs.

\subsection{Setup and Notation}
Let two networks $F$ and $G$ map an input $x \in \mathcal{X}$ to a hierarchy of hidden-layer activations. A probe dataset $\mathcal{D} = \{x_m\}_{m=1}^{M}$, typically composed of unlabeled validation or public test inputs, is used solely to elicit these activations. For each neuron $u$ in $F$, its activation vector is defined as
\(
\alpha_u = [h_u(x_1), h_u(x_2), \ldots, h_u(x_M)],
\)
where $h_u(x_m)$ denotes the scalar activation of neuron $u$ for input $x_m$. The same is done for all neurons in $G$. The similarity between two neurons’ activation vectors is measured by the Pearson correlation coefficient $\rho(\cdot, \cdot)$.

\subsection{Per-Neuron Best-Match Score}
For every neuron $u$ in $F$, we identify the neuron in $G$ whose activation pattern is most strongly correlated with it. This neuron, denoted $v^*(u)$, is found as
\(
v^*(u) = \arg\max_{v \in \mathcal{U}_G} |\rho(\alpha_u, \alpha_v)|,
\)
where $\mathcal{U}_G$ is the set of all hidden neurons in $G$. The absolute value accounts for possible sign inversions caused by subsequent linear transformations or normalization layers.

To incorporate architectural depth information, we introduce a layer-distance penalty. The function $\text{layer}(\cdot)$ returns the integer index of the layer that a neuron belongs to. In multilayer perceptrons (MLPs), layers are numbered sequentially starting from the first hidden layer. In convolutional networks (CNNs), each convolutional or pooling block can be indexed as a layer, and in hierarchical architectures such as ResNets, DenseNets, or EfficientNets, each stage or transition block may share a single index to represent comparable levels of abstraction.

The layer-distance-normalized correlation score for neuron $u$ is defined as
\[
S(u; F \rightarrow G) =
\frac{|\rho(\alpha_u, \alpha_{v^*(u)})|}
{1 + |\mathrm{layer}(u) - \mathrm{layer}(v^*(u))|}.
\]
The denominator penalizes matches between neurons that are far apart in depth, ensuring that correspondences between early- and late-layer neurons contribute less to the final measure. The same computation is performed symmetrically for each neuron $v$ in $G$, identifying its best match in $F$.

\subsection{Network-Level Correlation}
The overall cross-model neuronal correlation is obtained by averaging per-neuron scores from both directions. This ensures that the result is invariant to which model is considered the reference:
\[
\mathrm{Corr}(F, G) =
\frac{1}{2} \Bigg(
\frac{1}{|\mathcal{U}_F|}
\sum_{u \in \mathcal{U}_F} S(u; F \rightarrow G)
+
\frac{1}{|\mathcal{U}_G|}
\sum_{v \in \mathcal{U}_G} S(v; G \rightarrow F)
\Bigg).
\]
The resulting scalar lies in the interval $[0,1]$. Higher values indicate stronger representational similarity between networks, while lower values suggest that the two models have developed distinct internal structures.

\subsection{Partial Correlation for Tractability}
The computational complexity of full cross-layer matching is quadratic in the number of neurons, $\mathcal{O}(|\mathcal{U}_F||\mathcal{U}_G|)$, which is impractical for modern architectures with millions of activations. To make the computation feasible, we use a partial correlation strategy. The comparison is restricted to corresponding or functionally similar layers 
rather than all possible cross-layer pairs. Additionally, a random subset of neurons can be sampled from each layer to further reduce computational load.

This approximation substantially reduces cost while preserving the ability to detect meaningful alignment between two architectures’ representational spaces. Empirically, partial correlation values remain stable across reasonable subsampling levels and layer selection schemes.

\section{Empirical Results}

\begin{table}[h]
\caption{Partial correlation between ResNets.}
\label{table:resnet}
\centering
\resizebox{0.75\columnwidth}{!}{
\begin{tabular}{|c|c|c|c|c|c|}
\hline
     & ResNet-18 & ResNet-34 & ResNet-50 & ResNet-101 & ResNet-152 \\ \hline
ResNet-18 & 1    & 0.661 & 0.288 & 0.312 & 0.133 \\ \hline
ResNet-34 & X    & 1     & 0.480 & 0.402 & 0.258 \\ \hline
ResNet-50 & X    & X     & 1     & 0.206 & 0.113 \\ \hline
ResNet-101 & X    & X     & X     & 1     & 0.052 \\ \hline
ResNet-152 & X    & X     & X     & X     & 1     \\ \hline
\end{tabular}
}
\end{table}

The proposed correlation metric was investigated on large neural networks such as ResNet \cite{resnet}, DenseNet \cite{densenet}, and EfficientNet \cite{tan2019efficientnet}. 
For all cases, we utilized the pretrained weights on ImageNet \cite{imagenet} as our baselines. We limited our analysis to just 10 test data points sampled from the validation set of ImageNet due to the time and space complexities inherent in our methodology. Since investigating the correlation on all of the neurons in all layers was not possible, we focused our analysis solely on the initial output of the fourth layer for all ResNets, the primary output of the third transition layer for all DenseNets, and output of the third stage for all EfficientNets. A point to be highlighted is that assessing the correlation metric on the final layers often yields superior accuracy compared to other layers, given that these layers encapsulate more profound and meaningful representations. The results shown in Table \ref{table:resnet} suggest that the most similar networks to ResNet-18, ResNet-34, ResNet-50, ResNet-101, and ResNet-152 are ResNet-34, ResNet-18, ResNet-34, ResNet-34, and ResNet-34 respectively. 
As demonstrated in Table \ref{table:densenet}, DenseNet-161, DenseNet-121, DenseNet-201, and DenseNet-121 have the most similarities to DenseNet-121, DenseNet-161, DenseNet-169, and DenseNet-201. Similarly, Table \ref{table:efficientnet} shows high correlation between adjacent scales of EfficientNet.
This pattern illustrates that networks with roughly the same number of layers often exhibit more profound partial correlations, serving as a validation of the effectiveness of our proposed correlation metric. 

\begin{table}[t]
\caption{Partial correlation between DenseNets.}
\label{table:densenet}
\centering
\resizebox{0.75\columnwidth}{!}{
\begin{tabular}{|c|c|c|c|c|}
\hline
& DenseNet-121 & DenseNet-161 & DenseNet-169 & DenseNet-201 \\ \hline
DenseNet-121 & 1    & 0.780 & 0.718 & 0.766  \\ \hline
DenseNet-161 & X    & 1     & 0.697 & 0.748 \\ \hline
DenseNet-169 & X    & X     & 1     & 0.725 \\ \hline
DenseNet-201 & X    & X     & X     & 1     \\ \hline
\end{tabular}
}
\end{table}

\begin{table}[t]
\caption{Partial correlation between EfficientNets.}
\label{table:efficientnet}
\centering
\resizebox{0.95\columnwidth}{!}{
\begin{tabular}{|c|c|c|c|c|c|}
\hline
& EfficientNet-B0 & EfficientNet-B1 & EfficientNet-B2 & EfficientNet-B3 & EfficientNet-B4 \\ \hline
EfficientNet-B0 & 1       & 0.820 & 0.811 & 0.816 & 0.819 \\ \hline
EfficientNet-B1 & X       & 1      & 0.826 & 0.820 & 0.821 \\ \hline
EfficientNet-B2 & X       & X      & 1      & 0.819 & 0.814 \\ \hline
EfficientNet-B3 & X       & X      & X      & 1      & 0.822 \\ \hline
EfficientNet-B4 & X       & X      & X      & X      & 1      \\ \hline
\end{tabular}
}
\end{table}

\section{Discussion}

These experimental results indicate that our proposed correlation score can effectively demonstrate connections and relationships across diverse neural networks. While these findings are still preliminary, we assert that this score can potentially offer insights into various factors including:

\begin{itemize}
\item \textbf{Architectural size impact:} Shallower architectures tend to exhibit higher correlation with other networks due to their simplicity and shared properties with most neural networks. 
A greater number of neurons per layer can contribute to a higher correlation score, as it allows for the recognition of a broader range of similar patterns with increased neural capacity.
\item \textbf{Layer depth impact:}
Calculating partial correlations  can yield more interpretable insights. 
Evaluation of the partial correlation metric on the earlier layers is typically less accurate since the final layers tend to capture more sophisticated and significant representations of the data.
\item \textbf{Performance implications:} Lower correlations may correlate with poorer model performance, indicating that neuron activations deviate from the expected patterns necessary for accurate decision-making.
\end{itemize}

We further note that despite the valuable insights provided by the proposed correlation method, a principle drawback lies in the suboptimal performance in terms of time complexity for obtaining this score. Moreover, when dealing with larger models, a more efficient approach for calculating correlation with improved time complexity and precision may be necessary. Another limitation of our current methodology is the inability to pinpoint the exact reasons behind a low correlation score. However, if a high correlation is observed between two networks, where one is recognized for its accuracy and robustness, there is a strong indication that the other network is also performing well and likely to generalize. Nevertheless, this inference is not entirely precise.

\clearpage

\bibliography{iclr2026_conference}
\bibliographystyle{iclr2026_conference}


\end{document}